\documentclass[letterpaper, 10 pt, conference]{ieeeconf}  

\IEEEoverridecommandlockouts                              

\usepackage{amsmath} 
\usepackage{xcolor}
\usepackage{graphicx}
\usepackage{subfig}
\usepackage{type1cm}
\usepackage{lettrine} 
\usepackage[pagebackref=true]{hyperref}

\title{\LARGE \bf
Pouring Dynamics Estimation Using Gated Recurrent Units
}

\author{Qi Zheng
}

\begin{document}
\maketitle
\thispagestyle{empty}
\pagestyle{empty}

\begin{abstract}
One of the most commonly performed manipulation in a human’s daily life is pouring. Many factors have an effect on target accuracy, including pouring velocity, rotation angle, geometric of the source, and the receiving containers. This paper presents an approach to increase the repeatability and accuracy of the robotic manipulator by estimating the change in the amount of water of the pouring cup to a sequence of pouring actions using multiple layers of the deep recurrent neural network, especially gated recurrent units (GRU). The proposed GRU model achieved a validation mean squared error as low as $1e-4$ (lbf) for the predicted value of weight f(t). This paper contains a comprehensive evaluation and analysis of numerous experiments with various designs of recurrent neural networks and hyperparameters fine-tuning.
\end{abstract}

\section{Introduction}

\lettrine{R}obotics has positively transformed human lives and work, raise efficiency and safety as well as provide enhanced services. Especially in the restaurant and cooking environment, where safety and sanitation are essential, the cooking robot comes in handy. To ensure the robot can achieve human accuracy in performing the motions and have the ability to adjust according to the environment, such as cooking materials and object states, much work has been done on manipulation motion taxonomy \cite{c2}, different grasp taxonomies, and grasp types \cite{grasp1,grasp2,grasp3,grasp4} to allow a robot to "understand" and execute the proper motion.

Not only motion taxonomy, but cooking robots are also required to have detailed knowledge on various manipulation tasks in order to successfully perform kitchen activities, and accuracy has a key effect on the resultant product. Based on the Functional Object-Oriented Network (FOON) video set, pick-and-place is the most frequently executed motion, and pouring is the second most \cite{motionF1,motionF2}. For pick-and-place, researchers have used visual perception on a robot-arm system for carrying out flexible pick and place behavior \cite{pap2} and used a task-level robot system to carried out dozens of such operations that involving various complex environments \cite{pap1}. For this project, we focused on the pouring motion, several researchers have examined different pouring factors and approaches to increase the robot pouring skill and accuracy. For instances, using stereo vision to recognize liquid and particle flow during pouring \cite{pour1}, applying two-degrees-of-freedom-control to control the liquid level \cite{pour2}, using parametric hidden Markov models to allow force-based robot learning pouring skills \cite{pour3}, and using an RGB-D camera to track the liquid level \cite{pour4}. 

Besides using vision methodology, recurrent neural network (RNN) has been increasing in popularity for sequence learning and generation. Several studies used RNN to model liquid behavior \cite{rnn_pour1} and model pouring behavior\cite{rnn_pour2,rnn_pour3,rnn_pour4}.

\begin{figure}[ht]
  \centering
  \includegraphics[width=8cm]{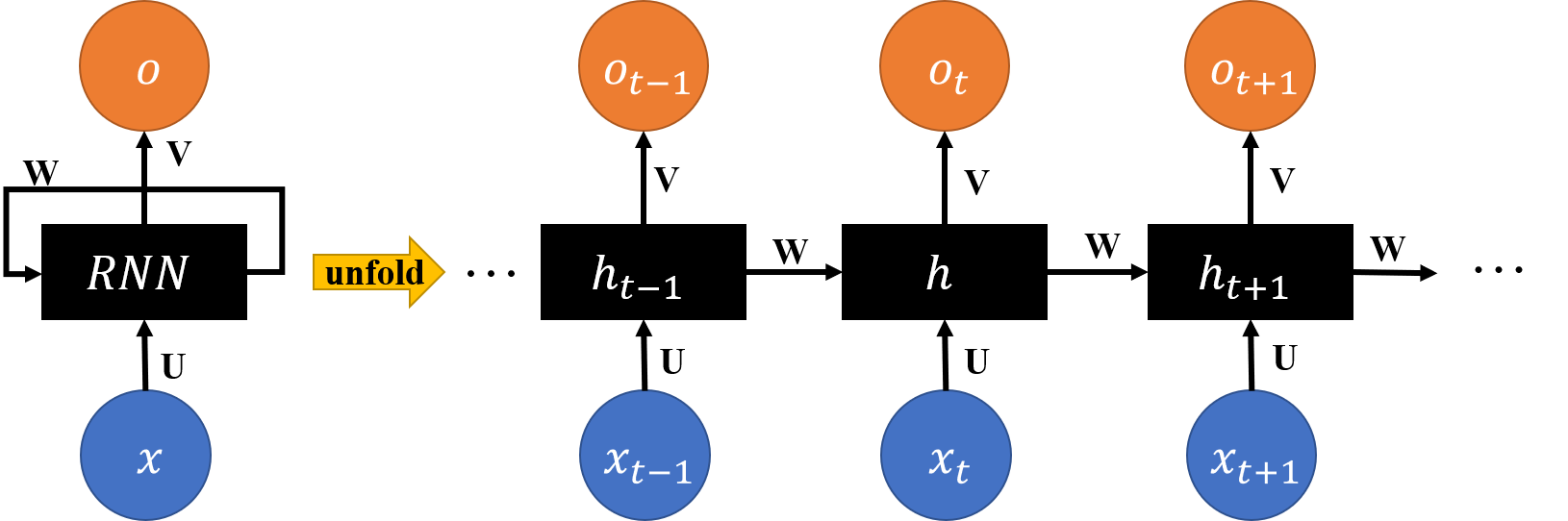}
  \caption{Recurrent Neutral Network}
  \label{fig:rnn}
\end{figure}

RNN is a type of artificial neural network that inherently suitable for sequential data or time-series data. In RNN, the connection of the units is formed as a directed graph along the sequence, which allows it to exhibit dynamic temporal behavior for the time sequence. The core idea of RNN is to use the information from the previous time step in the sequence to produce the current output, and the process will continue until all sequences are given input. Unlike other neural networks, in RNN, all the inputs are related to each other. Fig. \ref{fig:rnn} gives an illustration. At the last step, RNN has all the information from the previous sequences to produce predictive results. There are various types of RNN and each has its advantages and disadvantages. Couple studies on pouring motion used Peehole Long Short-Term Memory \cite{rnn_pour2, rnn_pour3,rnn_pour4}. Hence, for this project, three other common types of recurrent neural networks, simple RNN, LSTM, and Gated Recurrent Units (GRU), are experimented with to model the pouring behavior. Mechanisms are explained in section \ref{simpleRNN}.

\section{Data and Preprocessing}

The dataset contains a total of 688 pouring sequences and their corresponding weight measurements. Each motion sequence has seven feature dimensions, which for each timestamp of a motion sequence are

\hspace{-0.5cm}[
\begin{align*}
&\theta (t)   \hspace{0.4cm}  \text{  Rotation angle at time $t$ ($degree$)}\\
&f(t) \hspace{0.35cm} \text{   Weight at time $t$ ($lbf$)}\\
&f_{init} \hspace{0.3cm} \text{ Weight before pouring ($lbf$)}\\
&f_{target} \hspace{0.0cm} \text{   Weight aimed to be poured in the receiving cup ($lbf$)}\\
&h_{cup} \hspace{0.35cm} \text{ Height of the pouring cup ($mm$)}\\
&d_{cup} \hspace{0.35cm} \text{ Diameter of the pouring cup ($mm$)}\\
&\dot{\theta}(t) \hspace{0.4cm} \text{ Velocity of the pouring cup ($rad/s$)}
\end{align*}
]

\begin{figure}[ht]
  \centering
  \includegraphics[width=8cm]{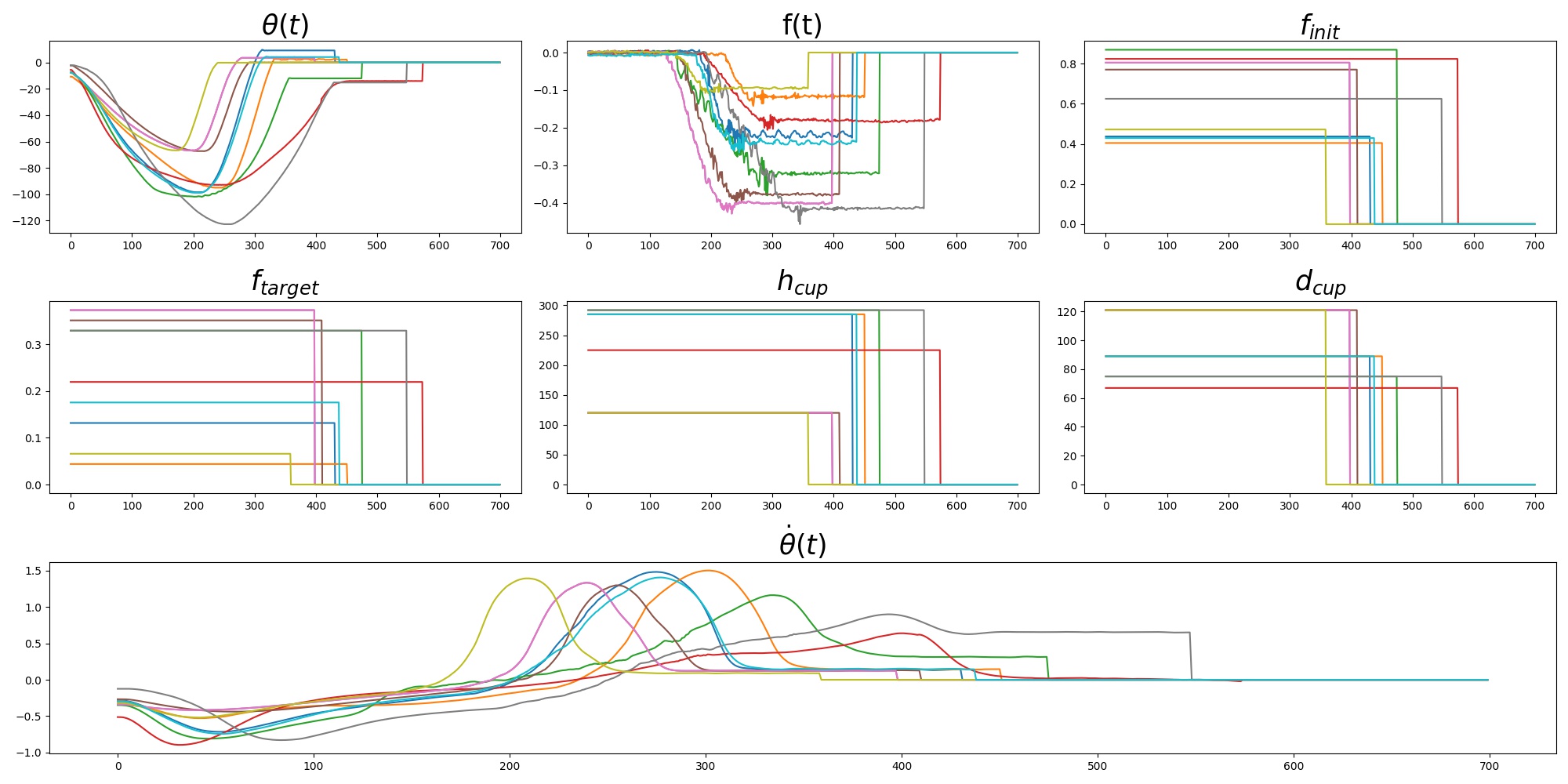}
  \caption{Dataset consists of seven features in each sequence}
  \label{fig:given_dataset}
\end{figure}

Only $\theta$, $f(t)$ and $\dot{\theta}(t)$ are changing with time, the other four sequences are constant throughout the entire sequence. The length of sequences is various, as all sequences are padded with zeros at the end according to the maximum length of the sequence, which in our case is 700. The purpose of zero post-padding is so that all sequences in a batch can fit in a standard length. Masking is used during the training to exclude padded zeros when computing the loss. Fig. \ref{fig:given_dataset} gives a simple illustration of given dataset. The detailed data collection process could be found in \cite{data}.

\subsection{Data Preprocessing}
Input $x(t)$ of the network contains a total of six features, with output of the network being one dimension $y(t)$,
\begin{align}
    x(t)&=[\theta (t), f_{init},f_{target},h_{cup},d_{cup},\dot{\theta}(t)]\\
    y(t)&=f(t)
\end{align}

Although, features $f_{init}$, $f_{target}$, $h_{cup}$, $d_{cup}$ stay constant throughout the time, those features still have affect on the target $f(t)$ and this notion has been proved in several experiments on different combination of the input features.

Before feeding the data into a neural network, input features $x(t)$ are being normalized to speed up the learning process which leads to faster convergence. Two common data normalization methods are min-max normalization and standardization.

\begin{figure}[ht]
  \centering
  \includegraphics[width=8cm]{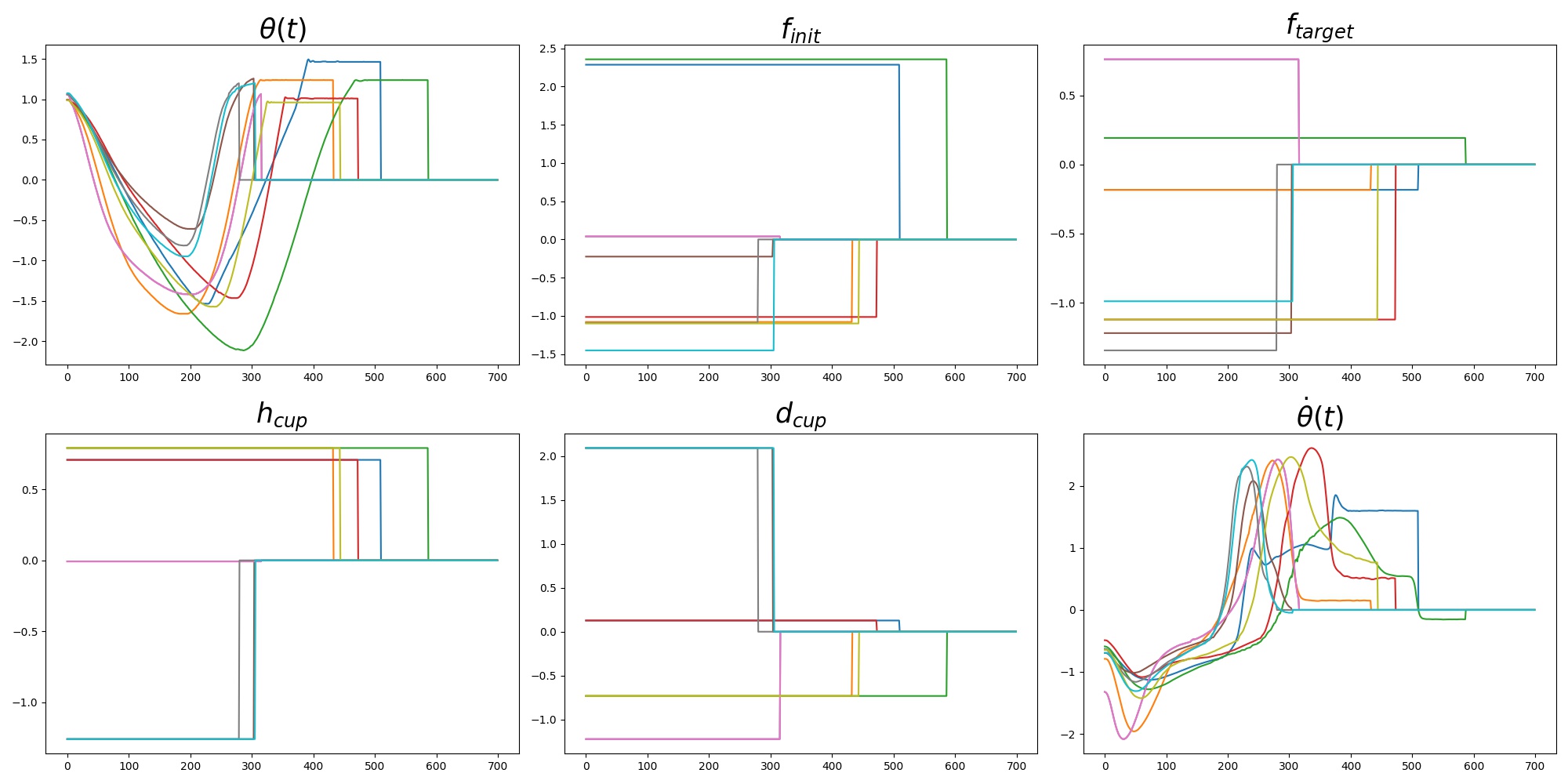}
  \caption{Normalized Input Features}
  \label{fig:scaled_dataset}
\end{figure}

Min-max normalization retains the original distribution of values except for a scaling factor and then transforms all the values to the common range of 0 and 1. However, this technique is not robust due to the high sensitivity to outliers and uncertainty of features of the test set. Therefore, a standard scale is being used to normalize the input features. The input features are being standardized independently on each feature by removing the mean and scaling to unit variance, the formula is shown in Eq. \ref{eq:z_forumla}, where $\mu$ is the mean of the training sample, and $s$ is the standard deviation of the training samples.
\begin{align} \label{eq:z_forumla}
z=\frac{(x-\mu)}{s}
\end{align}
Only real data are being normalized whereas zero-padding remind the same. Fig. \ref{fig:scaled_dataset} gives a simple illustration of standardized input data.

The data are being shuffled and randomly split into 80\%, 550 trials, for training and 20\%, 138 trials, for validation. The test dataset is reserved by the TA and instructor for model testing. The same normalization scale from the training dataset is being applied to the validation and test set.

\section{Methodology}

\subsection{RNN Architectures}

\textbf{Simple Recurrent Neural Network (Simple RNN)} \label{simpleRNN}
The simple RNN has a short-term memory problem due to vanishing and exploding gradient. To put it differently, simple RNN has a difficult time solving a problem that requires learning long-term temporal dependencies, therefore hampering learning of long data sequences as it processes more steps. The gradient is used to update the parameters in the network, and when the gradient becomes smaller and smaller, the parameter updates becomes insignificant which results in the network not learning from the earlier inputs. 

\begin{figure*}%
    \centering
    \subfloat[\centering Simple RNN cell]{{\includegraphics[width=5.5cm]{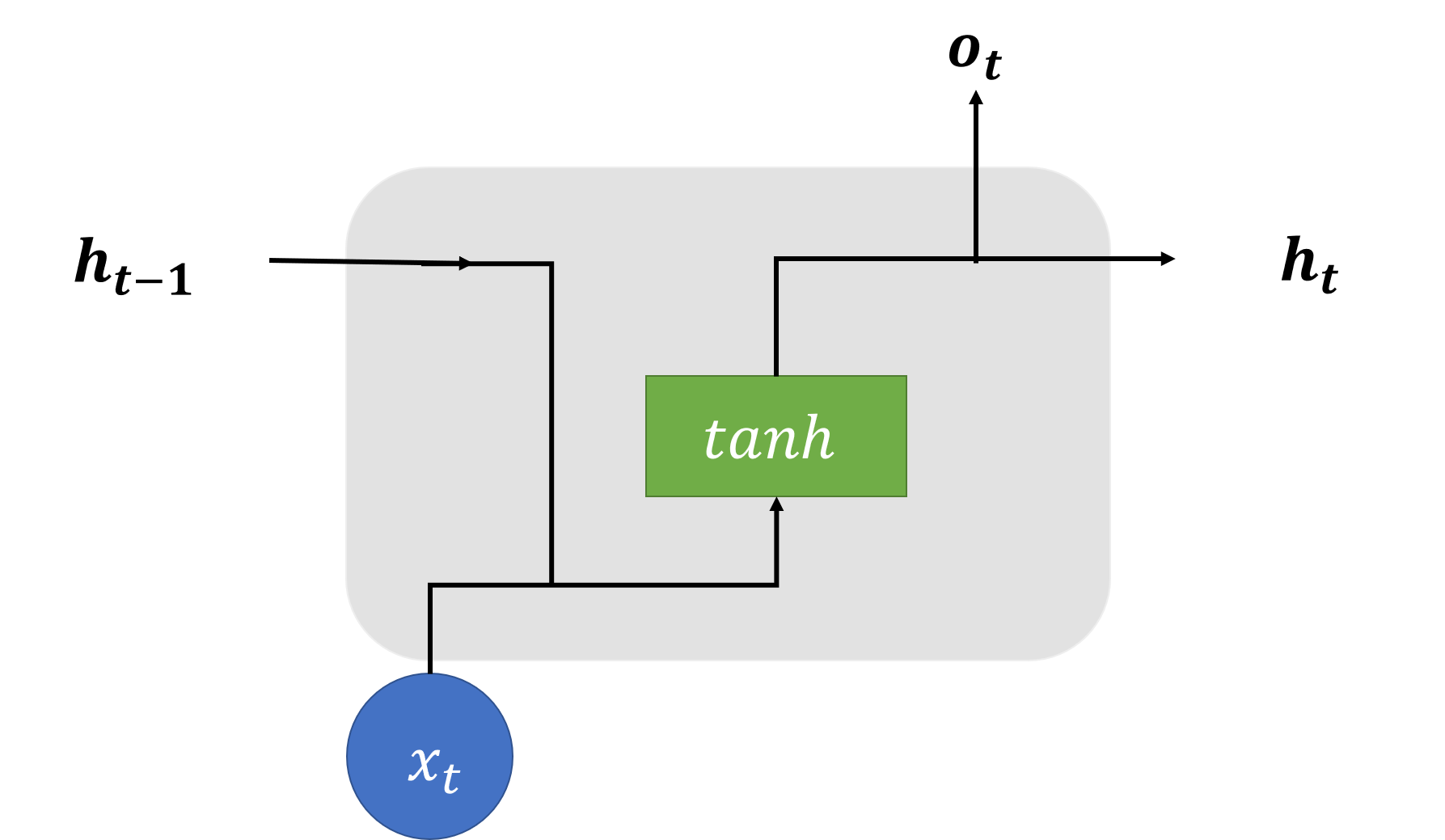} }}%
    \qquad
    \subfloat[\centering LSTM cell]{{\includegraphics[width=5cm]{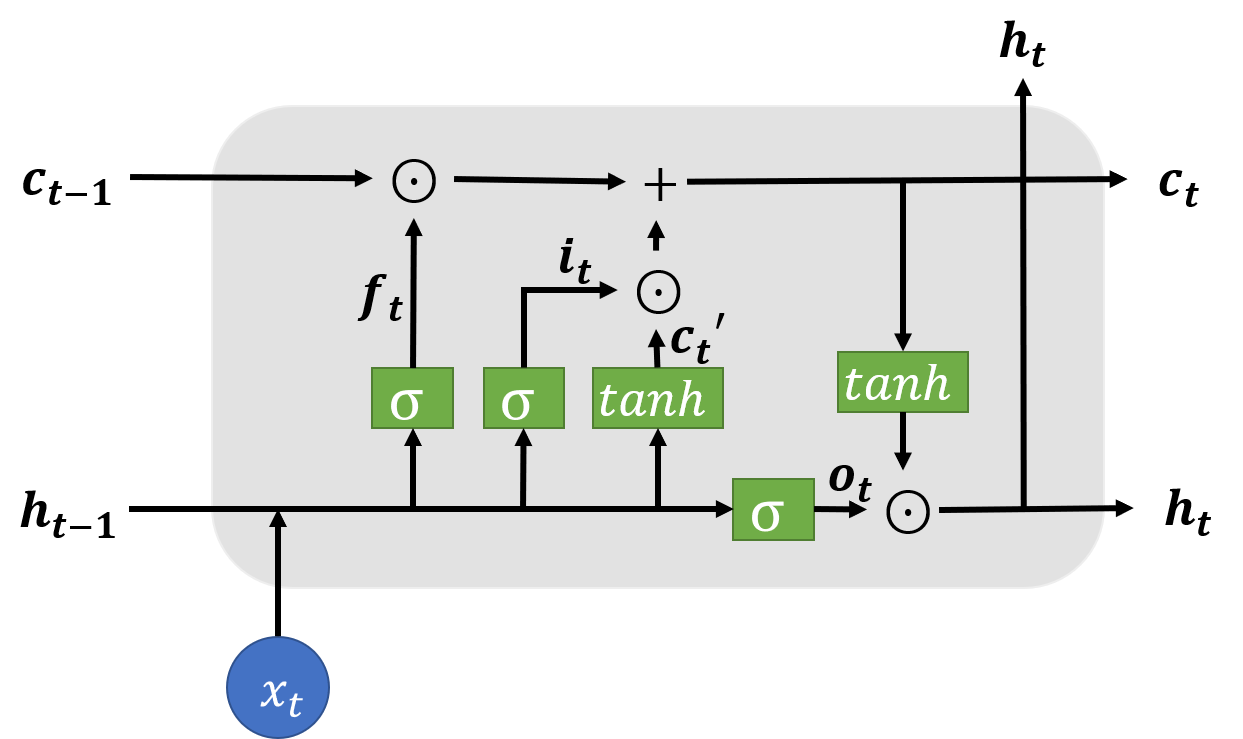} }}%
    \qquad
    \subfloat[\centering GRU cell]{{\includegraphics[width=5cm]{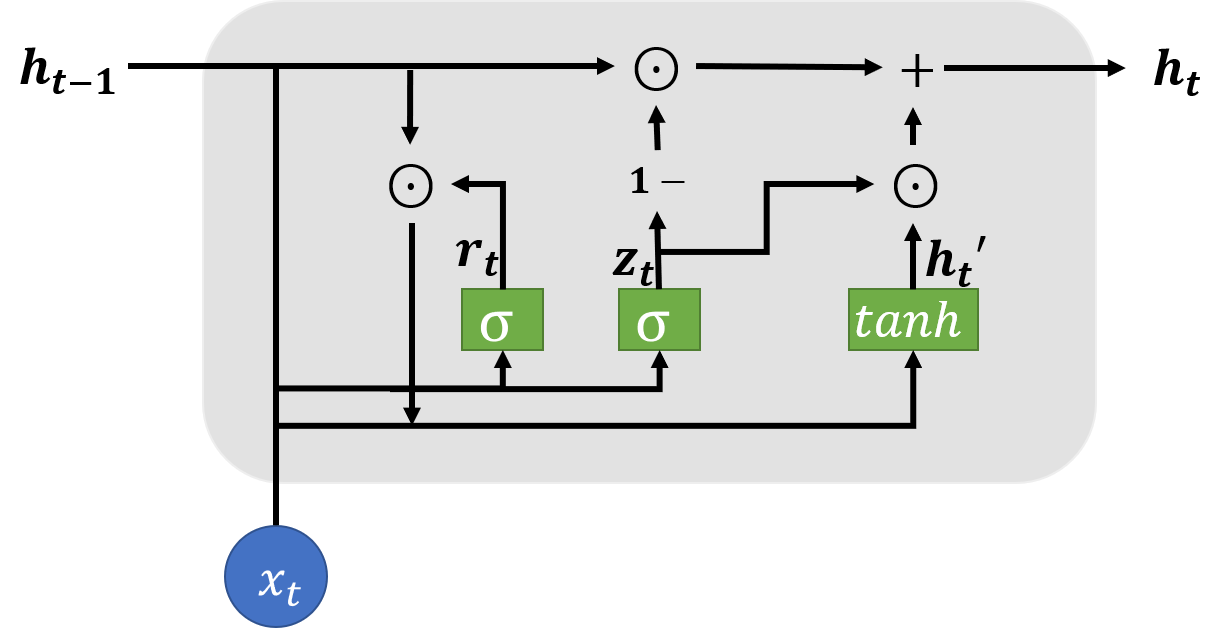} }}%
    \caption{Mechanism of three types of vanilla recurrent neural network cell}%
    \label{fig:rnn_model}%
\end{figure*}

The mechanism of simple RNN is illustrated in Fig. \ref{fig:rnn_model} (a) and is written as:
\begin{align}
    h_t &=\tanh_h(W_hx_t+W_hh_{t-1}+b_h)\\
    y_t&=\tanh_y(W_yh_t+b_y)
\end{align}
where $x_t$ is the given input vector, $y_t$ is the output vector, $h_{t-1}$ is output from the previous step, $h_t$ is the output at the current step, and $W$ is the weight parameter.

\textbf{Long Short-Term Memory (LSTM) } \label{LSTM}
LSTM is one of the most popular RNN that overcomes the vanishing gradient problem in back-propagation \cite{c1}. The mechanism of LSTM is illustrated in Fig. \ref{fig:rnn_model} (b) and is written as:
\begin{align}
    i_t &= \sigma(W_i [h_{t-1}, x_t]+b_i) \label{lstm_it}\\
    f_t &= \sigma(W_f [h_{t-1}, x_t]+b_f)\label{lstm_ft}\\
    o_t &= \sigma(W_o [h_{t-1}, x_t]+b_o)\\
    {c_t}' & = \tanh(W_c[h_{t-1}, x_t]+b_c)\\
    c_t& = f_t\odot c_{t-1}+i_t \odot {c_t}'\\
    h_t& =o_t \odot tanh(c_t)
\end{align}
where $i_t$, $f_t$, $o_t$ are the input, output, and forget gates respectively, $c_t$ is the cell state, $c'$ is the candidate cell, $\sigma$ is the sigmoid activation function, and $\odot$ represents the pointwise multiplication. 

LSTM contains both cell states and hidden states, where the cell state has the ability to remove or add information to the cell and maintain the information in memory for long periods of time. The introduction of gating mechanism in LSTM
\begin{itemize}
\item Input gate $i$: Update the cell status.
\item Forgot gate $f$: Decides how much information from the previous state should be kept and what information can be forgotten.
\item Output gate $o$: Determines the value for the next hidden state, which contains information o previous inputs.  
\end{itemize}
allows better control over gradient flow and better preservation of long-term dependencies.

\textbf{Gated Recurrent Units (GRU)} \label{GRU}
GRU is another popular RNN that is intended to solve the vanishing gradient problem. GRU contains only two gates, reset gate and update gate, and it is less complex than LSTM for that reason. The mechanism of GRU is illustrated in Fig. \ref{fig:rnn_model} (c) and is written as:
\begin{align}
    z_t &= \sigma(W_z [h_{t-1}, x_t]) \label{GRU_zt}\\
    r_t &= \sigma(W_r [h_{t-1}, x_t]) \label{GRU_rt}\\
    {h_t}' &= \tanh(W+h[r_t\odot h_{t-1}, x_t]+b_h)\\
    h_t&=(1-z_t)\odot h_{t-1}+z_t\odot {h_t}'
\end{align}
GRU shares many common properties with LSTM, where gating mechanism is also used to control the memorization process. GRU contains two gates, which are
\begin{itemize}
\item Update gate $z$: Decides whether the cell state should be updated with the candidate state.
\item Reset gate $r$: Decides whether the previous cell state is important or not.
\end{itemize}

By comparing GRU with LSTM, one can observe that GRU [Equ. \ref{GRU_zt}-\ref{GRU_rt}] is similar to LSTM [Equ. \ref{lstm_it}-\ref{lstm_ft}]. However, GRU requires less memory, is significantly faster to compute than LSTM due to GRU, uses fewer training parameters, and uses fewer gates.

\subsection{Proposed Architecture }
\begin{figure*}[ht]
  \centering
  \includegraphics[width=\linewidth]{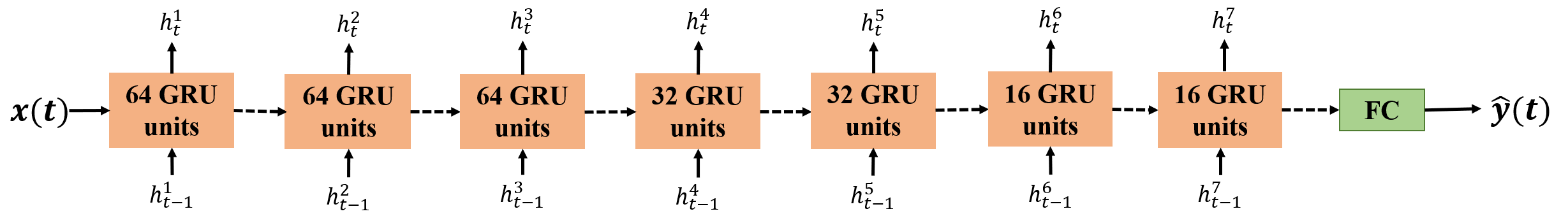}
  \caption{Proposed GRU Architecture Consists of Seven GRU layers}
  \label{fig:proposed_architecture}
\end{figure*}

The initial model experimented with was a proposed architecture by \cite{rnn_pour2}, which consists of a total of 4 layers, with each layer including 16 LSTM cells. After trying and training different recurrent neural networks with a varying number of layers and units, GRU gave better results than simple RNN and LSTM. The final model architecture is composed of a total of seven GRU layers and one fully connected layer, where each of the GRU layers returns a full sequence due to the following GRU layer needing a full sequence as the input. The mechanism of the GRU unit is explained in Section \ref{GRU}. Visualization of the architecture can be found in Fig. \ref{fig:proposed_architecture}. 

The resultant model contains a total of 83,537 parameters and they are all trainable parameters. The first seven layers of the networks are GRU layers, with three layers of 64 GRU cells, two layers of 32 GRU cells, and two layers of 16 GRU cells. Sigmoid activation is applied on each of the gates, update and reset gates, that is present in GRU, where value is in the range of 0 and 1. It is important to update and forgot data because any value multiplied by 0 is 0, which allows this data to be "forgotten", and any value multiplied by 1 is the value itself, allowing those data to be "kept". Therefore, the sigmoid function allows the network to learn the necessary information only. $\tanh$ is used as activation and is commonly used in RNN to overcome the vanishing gradient problem, where a function whose second derivative can sustain for a long-range before going to zero is need. $\tanh$ function squishes the values between -1 and 1 to regulate the output of the neural network.

The last layer in the network is a  fully connected layer, which reduces the output dimension to one. Dropout is not used and detailed analysis is explained in Section \ref{dropsection}.

\subsection{Loss Function}
For regression problems, the Mean Squared Error (MSE) is commonly used as a loss function for evaluating the performance. MSE is the mean overseen data of the squared differences between true and predicted values. The squaring is critical to reducing the complexity with the negative signs. MSE is defined as:
\begin{align}
    L=\frac{1}{n}\sum_{n=1}^{n}(y_i-\hat{y_i})^2
\end{align}
where $n$ is a number of data points, $y_i$ is observed values, and $\hat{y}$ is predicted values. 

Other loss functions such as Root Mean Squared Error (RMSE) and Mean Absolute Error (MAE) are also applied to the proposed models during the training, but neither perform better than MSE when compared the result in the same metric. The metric that used for this paper is MSE.

\subsection{Model Setting}
The best setting for the model is listed in Table \ref{final_hyper}. The model with the lowest validation loss was selected as the best pouring dynamics estimation model.

\begin{table}[ht]
\caption{Best Model Setting}
\label{final_hyper}
\begin{center}
\begin{tabular}{|c|c|}
\hline
Parameters & Setting\\
\hline
\hline
Optimizer & adam ($\beta_1 = 0.9$, $\beta_2 = 0.999$, $\epsilon = 10^{-9}$)\\
\hline
Batch Size & Default 32\\
\hline
Initial Learning Rate & $1e-3$\\
\hline
Learning Schedule & Constant Learning Rate\\
\hline
Number of epochs & 1500 epochs\\
\hline
\end{tabular}
\end{center}
\end{table}
The project is implemented using Keras and TensorFlow.


\section{Evaluation and Results}

\subsection{Simple RNN, LSTM, and GRU}
To determine the best RNN for the pouring dataset, the initial architecture is trained on the simple RNN, LSTM, and GRU individually for 500 epochs. The training is done using the Adam optimizer with a learning rate of $1e-3$. 

\begin{figure}[ht]
  \includegraphics[width=8cm, height = 6.5cm]{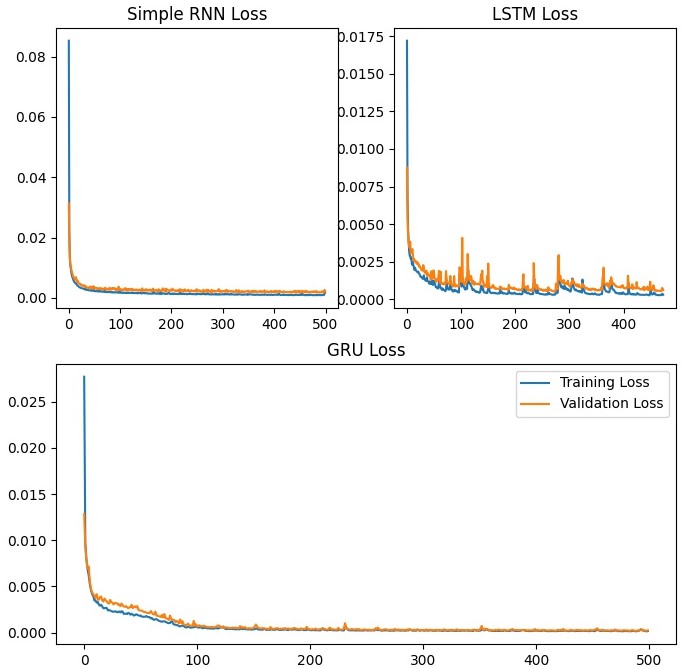}
  \caption{Model loss on simple RNN, LSTM, and GRU on 4 layers with 16 units each}
  \label{fig:diff_rnns}
\end{figure}

As a result, the initial architecture achieves an MSE of $1.74e-3$ and $5.09e-4$ on simple RNN and LSTM respectively, whereas GRU resulted in the best error rate of $1.97e-4$. From three loss graphs shown in Fig. \ref{fig:diff_rnns}, it is observed that loss for simple RNN converged right after the first 20 epochs and then stop decreasing, where a loss for LSTM was frequently fluctuating and unstable. Finally, GRU has a slightly better loss graph, where validation loss decreased continuously on the first 100 epoch but stabilized right after. Aside from error rates, GRU also has a faster training speed compared to simple RNN, which is much slower when training on this architecture. 

GRU has been shown to exhibit better performance on smaller and less frequent datasets than LSTM, where LSTM surpasses GRU on larger datasets \cite{perform}. In our case, only 550 sequences are used for training, which is quite small for the deep neural network. More experiments on GRU and LSTM are conducted in Section \ref{cellLayer}.




\subsection{Number of Cells and Layers} \label{cellLayer}
Many of the studies used a constant amount of internal cells throughout the RNN layers and keep the number layers within 1 to 4 \cite{rnn_pour1, rnn_pour2,rnn_pour3,rnn_pour4}. To examine the effect of the multi-layer network that contains different amounts of cells, several experiments are conducted by increasing or reducing the cells as it goes down the RNN layers.

From Table \ref{layerT}, it is observed that increasing the number of cells (Design 1-3) and the number of layers (Design 4-8) does yield a positive impact on the results, especially when using the GRU. Although the improvement is not significant, it shows that model complexity has an effect on the results. By comparing the loss value between GRU and LSTM, we see that GRU has a better performance in various architecture designs, except for Design 5 where the number of cells is increasing as it goes down the layers. More studies are needed to prove this notion.

Design 8, the proposed architecture, is the most complex of the architectures and it resulted in the lowest loss when using GRU, although LSTM did not benefit when adding the additional layer with 64 units from Design 7. Based on the observation, LSTM is frequently fluctuating, which could be the reason causing the network to perform the worst within the fixed epochs.

As mentioned, training on the same architecture, GRU is significantly faster than LSTM and LSTM is less stable than GRU. Those notions are again proved in the experiments. Therefore, consider both performance and computational cost, GRU is concluded as the best recurrent neural network for this application. 

\begin{table*}[ht]
\caption{Summary GRU and LSTM with Various Architectures (in MSE)}
\label{layerT}
\begin{center}
\begin{tabular}{|c|c|c|c|c|c|c|}
\hline
Design &Number of layers& Number of cells  at each layer& GRU training loss  & GRU validation loss & LSTM training loss & LSTM validation loss \\
\hline
\hline
1&1 & 16& $3.12e-4$ &$3.46e-4$& $5.17e-4$ &$8.62e-4$\\
\hline
2&1 & 32& $1.72e-4$&$2.36e-4$& $3.12e-4$ &$3.46e-4$\\
\hline
3&1& 64& $1.54e-4$&$2.15e-4$& $4.50e-4$ &$1.11e-3$\\
\hline
4&4& 16, 16, 16, 16 & $1.26e-4$&$1.97e-4$& $2.92e-4$&$5.09e-4$\\
\hline
5&4 & 8, 16, 32, 64&$1.4e-3$ &$2.1e-3$&$1.58e-4$&$2.56e-4$\\
\hline
6&4 & 64, 32, 16, 8 & $8.55-5$ &$1.81e-4$&$2.52e-4$&$4.09e-4$\\
\hline
7&6 & 64, 64, 32, 32, 16, 16&$8.87e-5$&$1.85e-4$&$1.24e-4$&$2.14e-4$\\
\hline
8&7 &64, 64, 64, 32, 32, 16, 16 &$7.18e-5$&$1.72e-4$&$3.54e-4$&$4.83e-4$\\
\hline
\end{tabular}
\end{center}
\end{table*}


\subsection{Optimizers}

Optimizer plays an important role in the neural network. A good optimizer can significantly reduce the loss and provide the most accurate results. By far, the most popular algorithms to perform optimization is SGD and Adam. In this experiment optimizers Adadelta, Adagrad, Adamax, and RMSprop have experimented. Training are done on a large number of epochs, 1000, in case some optimizers present a slower convergence. Default settings for each optimizer, see Table \ref{table_example}, are used with a learning rate of $1e-3$. 

\begin{table}[ht]
\caption{Optimizer Settings}
\label{table_example}
\begin{center}
\begin{tabular}{|c|c|}
\hline
Optimizer & Parameters Setting\\
\hline
\hline
Adadelta & $\rho = 0.95$, $\epsilon = 10^{-9}$\\
\hline
Adagrad & $\epsilon = 10^{-8}$, initial accumulator value = $0.1$\\
\hline
Adam & $\beta_1 = 0.9$, $\beta_2 = 0.999$, $\epsilon = 10^{-9}$\\
\hline
Adamax & $\beta_1 = 0.9$, $\beta_2 = 0.999$\\
\hline
RMSprop & $\rho = 0.9$, momentum = $0.0$, $\epsilon = 10^{-8}$\\
\hline
SGD & momentum = $0.9$, nesterov = False\\
\hline
\end{tabular}
\end{center}
\end{table}

\begin{figure}[t]
  \centering
  \includegraphics[width=8.5cm, height=5cm]{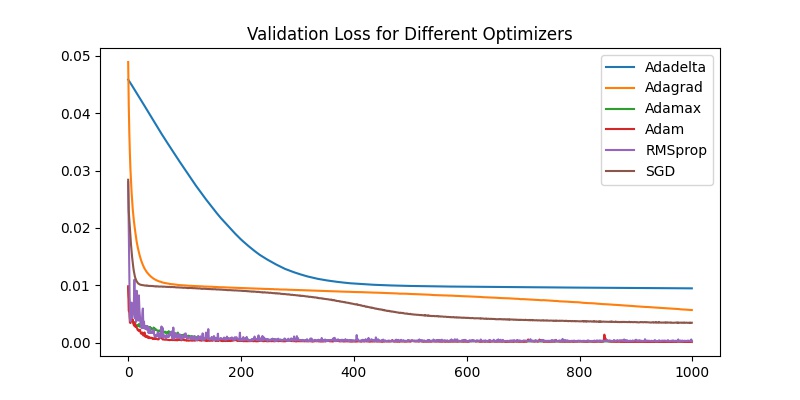}
  \caption{Validation Loss for Different Optimizers}
  \label{fig:optimizersLoss}
\end{figure}

From Fig. \ref{fig:optimizersLoss}, one can observe that Adam, Adamax, and RMSprop have relatively the same behavior and validation loss, although Adam and Adamax are much less fluctuating than RMSprop. SDG and Adagrad show an interesting trend, where validation loss has a visible decrease after several hundred epochs of unchanged, around 300 epochs for SGD and around 600 epochs for Adagrad. Compare with other optimizers, Adadelta's convergence rate is much slower and has a much higher validation loss. Of course, with different initial learning rates and decay schedules, some optimizers might have behaved better than others. However, based on the current setting, Adam is chosen as the best optimizer for the pouring dynamics RNN. 

\subsection{Learning Rate}
The learning rate controls how quickly a model is adapting to the problem. A smaller learning rate may allow the model to learn a more optimal set of weights, but it would require longer training time, whereas a larger learning rate will cause the model to converge too quickly to a suboptimal solution. Three common learning rate schedules have been experimented with: constant, step decay, and exponential decay. To observe the behavior of various changes on the learning rate, the same setting is applied to other hyperparameters. Model is being trained for 500 epochs and the lowest validation loss is being recorded in Table \ref{lr_exp}. 

One can observe that exponential decay performs worst out of three schedules while step decay performs better in general but does not take much advantage from lowering the learning rate. If the number of the epoch is increased, a smaller learning rate will cause very little to no updates to the weight in the network. Finally, a constant learning schedule with a rate of $1e-3$ surprisingly outperformed on both training and validation. Therefore, to allow the network to have a sustainable learning process, a constant learning rate $1e-3$ is used.
\begin{table}[ht]
\caption{Learning Rate Experiments Summary (in MSE)}
\label{lr_exp}
\begin{center}
\begin{tabular}{|c|c|c|c|}
\hline
Initial & Schedule & Training & Validation \\
 Learning Rate&  & Loss & Loss\\
\hline
\hline
$1e-1$ & constant &$2.18e-2$&$3.05e-2$\\
\hline
$1e-2$ & constant &$2.90e-4$&$2.24e-4$\\
\hline
$1e-3$ & constant &$1.02e-4$&$2.04e-4$\\
\hline
$1e-4$ & constant &$4.87e-4$&$7.23e-4$\\
\hline
$1e-2$ & exponential decay &$7.97e-4$&$1.15e-3$\\
\hline
$1e-3$ & exponential decay &$7.19e-3$&$7.59e-3$\\
\hline
$1e-2$ & step decay 0.5 every 10 epochs&$1.72e-4$&$3.14e-4$\\
\hline
$1e-2$ & step decay 0.5 every 20 epochs&$1.11e-4$&$2.17e-4$\\
\hline
$1e-2$ & step decay 0.5 every 30 epochs&$2.32e-4$&$4.18e-4$\\
\hline
$1e-2$ & step decay 0.5 every 40 epochs&$1.72e-4$&$3.56e-4$\\
\hline
\end{tabular}
\end{center}
\end{table}

\subsection{Dropout} \label{dropsection}
Dropout is commonly used to reduces overfitting. For RNN it is important not to apply dropout on the connection that conveys time related information \cite{dropout}. Experimentally, dropout rates 0.2, 0.1, and 0.05 were applied right after the last GRU and LSTM layer on the proposed architecture. As a result, the network performed worse on both training and validation sets. Therefore, no dropout is used in the final model. However, to avoid significant overfitting, the model is trained on the fixed epoch size, 1500 epochs, and the model with the lowest loss is kept. More importantly, GRU is less prone to overfitting since it only has two gates while LSTM has three, thus, dropout becomes less necessary for GRU.

\subsection{Result and Observation}
By training the model using the same setting multiple times, the proposed model is able to achieve MSE in the range of $9.29e-5$ to $1.97e-4$ on the validation dataset, which is relatively low and stable.

\begin{figure*}[ht]
  \includegraphics[width=\linewidth]{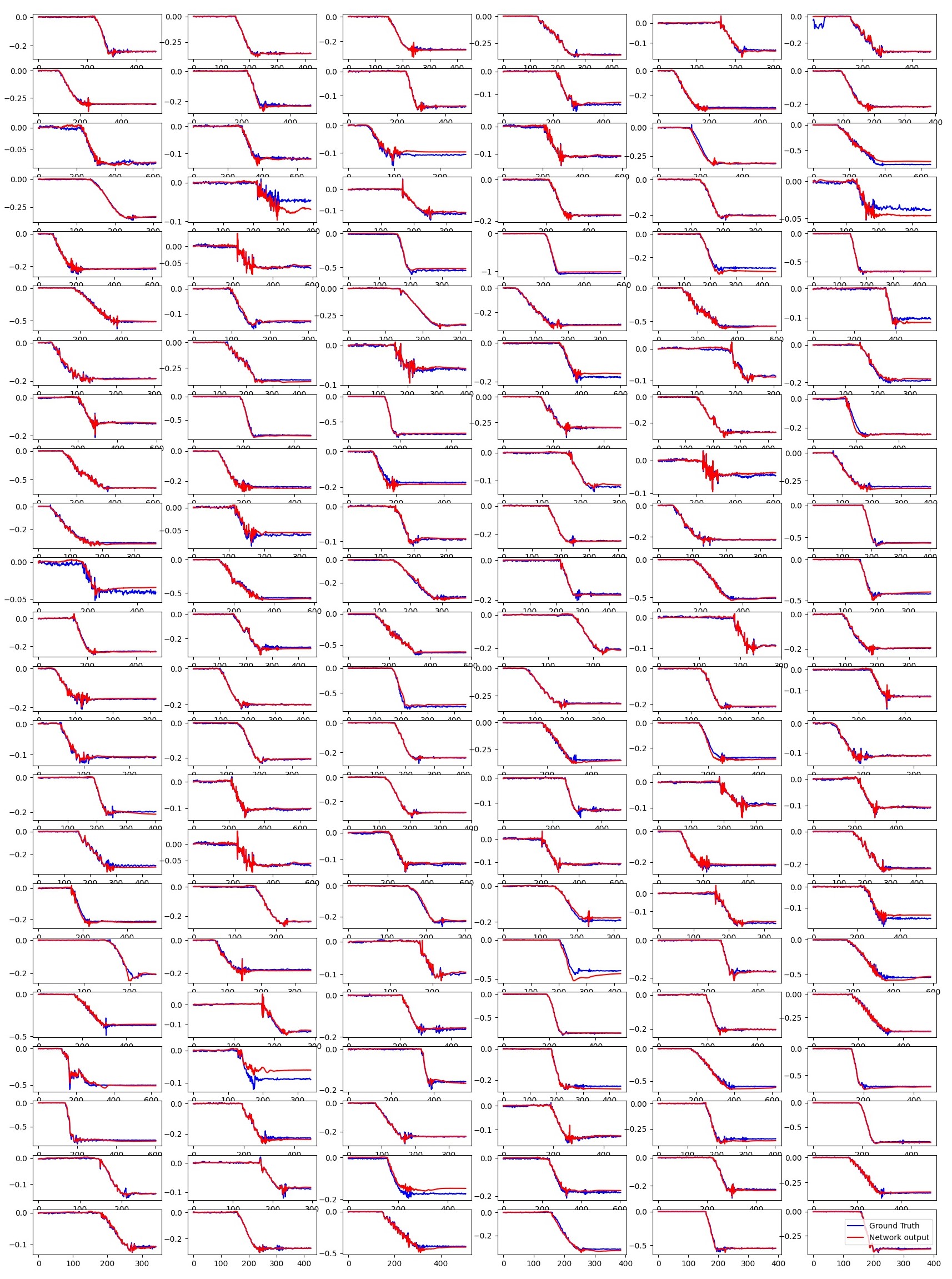}
  \caption{Ground Truth and Prediction Results on Validation Dataset}
  \label{fig:GroundTruthVSprediction}
\end{figure*}

The prediction and ground truth comparison graphs on validation dataset are shown in Fig. \ref{fig:GroundTruthVSprediction}, where x-axis is the time step $t$ and y-axis is the weight $f(t)$ with unit of $lbf$. Based on the observation, it can conclude that the model is able to learn and predict the general patterns, i.e. changing gradually in respect to time, in advance for the pouring motion. More importantly, the model is able to predict the change of the amount of water in the pouring cup accurately, where the blue line (ground truth) and the red line (network output) are close to perfectly match up. However, upon looking at the dataset, we observed that there is an outlier, possibly more, in the dataset (last sample in the first row of Fig. \ref{fig:GroundTruthVSprediction}), where the first 50 time steps show a small drop and rise in weight $f(t)$. This will cause the model to make an incorrect prediction on such a pattern around those time steps due to being unseen or rarely seen in the training dataset.

\section{CONCLUSIONS}

In this pouring dynamic estimation project, various recurrent neural networks have been investigated with different hyperparameter settings to estimate the change in the amount of water in the pouring cup to the sequences of pouring motion. Experimentally, it is found that GRU outperforms LSTM in both computational cost and performance based on the pouring dataset. In addition, the evidence shows that by training the exact same architecture using the same setting, the network prediction results in some variances, within the range of $1.041e-4$, even with no dropout applied.

The proposed model achieved an MSE as low as $9.29e-5$, which is a decent loss value. Based on the given dataset, it might reach its limitation already. However, there is some restriction in the experiment that may influence the potential of the model to achieve a lower error. The given dataset is a bit small for the deep neural network, hence, by increasing the number of trials, the model would be able to learn more features and patterns from the data. Additionally, the dataset should also contain more variants by increasing pouring data trials with different environment settings, materials, and other related features, such as rotation angle $\theta$, velocity $\dot{\theta}$ and various geometric of the cups. By doing so, the model can learn more variation directly from the dataset, which can result in a more robust network. There are plenty of researches going on for generating a dynamic response of motion sequences. The results of the robust model allow robotics to achieve human accuracy in executing the motions.

For future work, other types of RNN can experiment, such as continuous-time RNN (CTRNN), recurrent multi-layer perceptron network (RMLP), and multiple timescales RNN (MTRNN), etc. Different RNN architectures could result in different behaviors.



\end{document}